%% file: main.tex
\pdfoutput=1

\documentclass[11pt,a4paper]{article}
\usepackage[final]{eacl2023}
\usepackage{times}
\usepackage{latexsym}

\usepackage{hyperref}
\usepackage{url}
\usepackage{breakurl}

\usepackage{graphicx}
\usepackage[export]{adjustbox}
\usepackage{bm}
\usepackage{here}
\usepackage{amsmath}
\usepackage{amssymb}
\usepackage{amsfonts}
\usepackage{multirow}
\usepackage{arydshln}
\usepackage{enumitem}
\usepackage{dsfont}

\usepackage{algorithm}
\usepackage{algpseudocode}
\usepackage{soul}
\usepackage{float}
\usepackage{subcaption}
\usepackage{booktabs}
\title{DialogStudio: Towards Richest and Most Diverse Unified Dataset Collection for Conversational AI}
\author{
\centerline{Jianguo Zhang\thanks{\ \  Core contributors. Work completed during Kun’s internship at Salesforce. Zhiwei is also a major contributor.}~$^1$,~ Kun Qian\footnotemark[1]~$^2$,~ Zhiwei Liu$^1$, ~\textbf{Shelby Heinecke}$^1$, ~\textbf{Rui Meng}$^1$}  \\
\centerline{\textbf{Ye Liu}$^1$,  \textbf{Zhou Yu}$^2$, ~\textbf{Huan Wang}$^1$, ~\textbf{Silvio Savarese}$^1$, ~\textbf{Caiming Xiong}$^1$}\\
\centerline{$^1$ Salesforce AI ~ $^2$ Columbia University } \\
\centerline{jianguozhang@salesforce.com, kq2157@columbia.edu}}
\begin{document}

\maketitle

% \footnotetext[1]{Core contributors. Work completed during Kun's internship at Salesforce.}
% \footnotetext[2]{Also a major contributor.}

% \vspace{-1em}
\begin{abstract}
Despite advancements in conversational AI, language models encounter challenges to handle diverse conversational tasks, and existing dialogue dataset collections often lack diversity and comprehensiveness. To tackle these issues, we introduce DialogStudio: the largest and most diverse collection of dialogue datasets, unified under a consistent format while preserving their original information. Our collection encompasses data from open-domain dialogues, task-oriented dialogues, natural language understanding, conversational recommendation, dialogue summarization, and knowledge-grounded dialogues, making it an incredibly rich and diverse resource for dialogue research and model training.
To further enhance the utility of DialogStudio, we identify the licenses for each dataset, design external knowledge and domain-aware prompts for selected dialogues to facilitate instruction-aware fine-tuning. 
% Additionally, we identify dataset licenses, devise domain-aware prompts for selected dialogues to enable instruction-aware fine-tuning.
Furthermore, we develop conversational AI models using the dataset collection, and our experiments in both zero-shot and few-shot learning scenarios demonstrate the superiority of DialogStudio. 
To improve transparency and support dataset and task-based research, as well as language model pre-training, all datasets, licenses, codes, and models associated with DialogStudio are made publicly accessible\footnote{\url{https://github.com/salesforce/DialogStudio}}. 
\end{abstract}

% \footnote{Due to the extensive size ($\sim$50GB) of our data and code, they will be made publicly available upon the paper's publication.}

\input{1-introduction}

\input{2-data_analysis}
\input{4-datasets}
\input{6-exp_results}
\input{7-conclusion}

\clearpage
\bibliographystyle{acl_natbib}
% \bibliography{ref_custom, ref}
\bibliography{ref_neurips_2023}

\clearpage
\appendix

\begin{center}
\Large
\textbf{Appendix}
\end{center}

\input{8-appendix.tex}
% \input{Tex/appendix}
% \input{Tex/appendix_tod}

\end{document}

\maketitle

\setcounter{table}{5}
\setcounter{figure}{4}

\input{supplementary.tex}

\newpage
%\clearpage
\bibliographystyle{acl_natbib}
\bibliography{emnlp2020}

%% file: 1-introduction.tex
% \vspace{-.4em}
\section{Introduction}\label{sec:intro} 
% \vspace{-.4em}

% % Outlines
% % ------------------
% 1. current dialogue datasets are small and task-specific, in different format
% 2. there are some existing pre-trained dialogue model (cosmos, instructdial, godel), cosmos focuses on only open-domain dialogs, godel focuses on knowledge-grounded dialogs, instructdial focuses on only instruction tuning. no one includes all tasks.
% 3. we propose a collection of dialogue datasets, along with a general dialogue model.

% contribution:
%   1. collect and uniform 100+ datasets and publish it, fix the annotations of some datasets.
%   2. propose a general  dialogue model that can not only handle task-oriented dialogs, but also other dialogue tasks
%   3. integrate evaluation set for different dialogue tasks.
% % -----------------

% current dialogue datasets are small and task-specific, in different format
% There are some existing pre-trained dialogue model
% The existing pre-trained dialogue models, such as Cosmos~\citep{kim2022soda}, InstructDial~\citep{gupta2022instructdial}, and Godel~\citep{peng2022godel}, significantly contributes to the research in conversational AI.
% However, each of these models focuses on part of specific aspects of dialogue understanding and lacks the versatility to handle diverse conversational tasks.
% Cosmos primarily emphasizes open-domain dialogs.
% Godel focuses on knowledge-grounded dialogs.
% and InstructDial concentrates solely on instruction tuning.
% As a result, those models are still far from a comprehensive dialogue model that is able to effectively handle a wide range of dialogue tasks.

Recent years have seen remarkable progress in Conversational AI, primarily driven by the advent of approaches and language models~\citep{shuster2022blenderbot,zhang2023enhancing,longpre2023flan,touvron2023llama}. 
Despite the advancements, these models could fall short when handling various tasks in a conversation due to the lack of comprehensive and diverse training data. 
Current dialogue datasets~\citep{lin2021bitod,asri2017frames} are typically limited in size and task-specific, which thus results in suboptimal ability in task-oriented model performance.
Additionally, the lack of dataset standardization impedes model generalizability.

A few recent works~\citep{gupta2022instructdial,longpre2023flan,ding2023enhancing} have introduced a large collection of datasets, which includes diverse tasks based on public datasets. For instance, FlanT5~\citep{longpre2023flan} presents the flan collections with a wide array of datasets and tasks. Despite this breadth, the coverage of dialogue datasets within the Flan collection remains notably sparse, featuring only about ten datasets.   Although OPT~\citep{iyer2022opt} have incorporated collections with several dialogue datasets, these collections remain inaccessible to the public. In contract, efforts like InstructDial~\citep{gupta2022instructdial} and ParlAI~\citep{miller2017parlai} consist of more dialogue datasets, but they lack diversity and comprehensiveness. 
For instance, ParlAI mainly includes open-domain dialogue datasets, which are exclusively accessible through their platform. Other collections~\citep{gupta2022instructdial,kim2022soda,ding2023enhancing,dubois2023alpacafarm} often distill single dataset from ChatGPT or process datasets into a sequence-to-sequence format to support language model training, featuring only input-output pairs such as dialogue context and system response. However, previous collections often overlook other crucial dialogue information, constraining their utility for research  on individual datasets, tasks, and broader applications. 

% we introduce DialogStudio, the largest and unified collection of publicly available, task-specific dialogue datasets.

\begin{figure*}
    \centering
    \begin{subfigure}[b]{0.49\textwidth}
        \includegraphics[width=\linewidth]{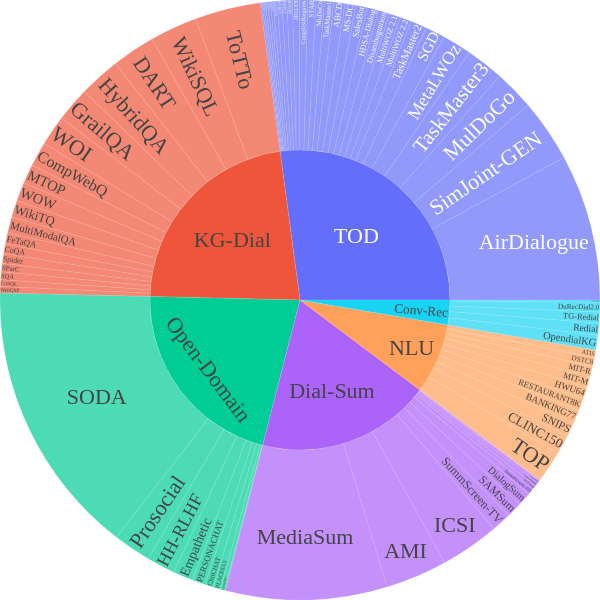}
        \caption{Dataset Distribution}\label{fig:stats_data}
    \end{subfigure}
    % \hspace{-2mm}
    \begin{subfigure}[b]{0.49\textwidth}
        \includegraphics[width=\linewidth]{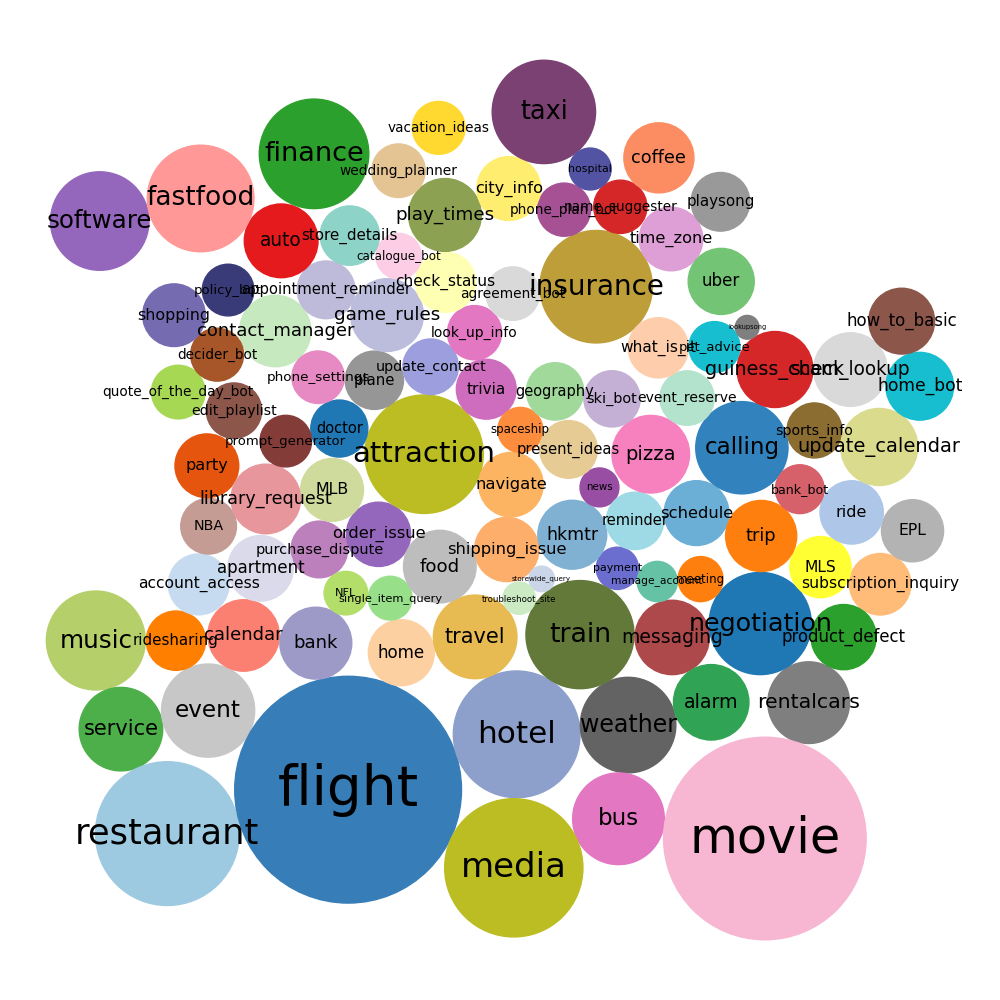}
        \caption{Domain Coverage of TOD}\label{fig:stats_domain}
    \end{subfigure}
\caption{(a) is the distribution of all datasets in DialogStudio. The outer and inner circle list names of datasets and the associated categories, respectively. (b) illustrates covered domains of Task-Oriented Dialogues in DialogStudio.}
\label{fig:overall}
\end{figure*}

To overcome the aforementioned challenges, we introduce DialogStudio, the most comprehensive and diverse collection of publicly available dialogue datasets, unified under a consistent format. By aggregating dialogues from various sources, DialogStudio promotes holistic analysis and the development of models adaptable to a variety of conversational scenarios. The collection spans an extensive range of domains, aspects, and tasks, and it is inclusive of several categories: Open-Domain Dialogues, Task-Oriented Dialogues, Natural Language Understanding, Conversational Recommendation, Dialogue Summarization, and Knowledge-Grounded Dialogues. Thus, it can provide support for research in both individual dialogue tasks and large-scale language pre-training.

% Furthermore, we identify specific dialogue domains for each dialogue and create tailored prompts accordingly. Utilizing selected datasets from DialogStudio, we develop a suite of instruction-aware models, known as DialogOhana, spanning from 220M to 3B in capacity. These models, capable of handling a variety of external knowledge, excel in both response generation and an assortment of general tasks.

DialogStudio stands out not only for its comprehensive coverage but also for its accessibility. It offers easy access with a unified format and documents.
A straightforward $\mathtt{load\_dataset()}$ command through HuggingFace allows users to seamlessly interact with the collection, and we have included documentation for each dataset to enhance usability. 
We anticipate that this collection will enable comprehensive and standardized training and evaluations of dialogue models, fostering fair comparisons and propelling further advancements in Conversational AI.

Furthermore, we identify dialogue domains, design external knowledge for available dialogues and create tailored prompts for selected datasets accordingly.  Leveraging these datasets from DialogStudio, we have constructed instruction-aware models, with capacities ranging from 770M to 3B parameters.  These models have the ability to handle various external knowledge and are adept at both response generation and general tasks, demonstrating the benefits of DialogStudio. The main contributions of this paper are as follows:
\begin{itemize}[leftmargin=*]
    \item We introduce DialogStudio, a meticulously curated collection of more than 80 dialogue datasets. These datasets are unified under a consistent format while retaining their original information. We integrate external knowledge, incorporate domain-aware prompts and identify dataset licenses, making DialogStudio an exceptionally rich and diverse resource for dialogue research and model training.
    \item We have made our datasets publicly available to enhance transparency and support research efforts. Additionally, we are committed to improving DialogStudio's usability and will persist in our efforts to refine it, ensuring an optimal user experience.
    % \item To promote transparency and facilitate research, we make our collected datasets publicly accessible.  Moreover, we have made concerted efforts to  enhance DialogStudio's usability and will continue refining it for optimal user experience.
    \item We train conversational AI models based on DialogStudio, and these models have demonstrated superior performance over strong baselines in both zero-shot and few-shot learning scenarios. 
\end{itemize}

% Moreover, we address inconsistencies and improve the annotations of certain datasets to enhance their quality and usability. 
% By consolidating these datasets, DialogStudio provides a qualitative resource for training and validating dialogue models across various tasks.
% The consolidation and standardization of diverse dialogue datasets within DialogStudio provide several advantages. Firstly, it allows researchers and practitioners to overcome the limitations imposed by individual datasets, fostering the development of more robust and generalizable conversational models. Secondly, the unified dataset format facilitates the exploration of novel research questions, benchmarking, and evaluation, enabling researchers to gain deeper insights into the performance and limitations of different models. 

% Furthermore, DialogStudio supports a unified collection of evaluation sets, covering task-oriented dialog, conversational QA, knowledge-grounded dialog, etc.
% Those evaluation datasets are independent of the training set, covering both seen and unseen domains. 

% \item We integrate evaluation sets tailored explicitly to different dialogue tasks.

%% file: 2-data_analysis.tex
\section{Data analysis}
% \textcolor{red}{I think we need some analysis for the datasets. People at least wants to know further dataset information and their qualities}

\begin{figure*}
    \centering
    \begin{subfigure}[b]{0.32\linewidth}
        \includegraphics[width=\linewidth]{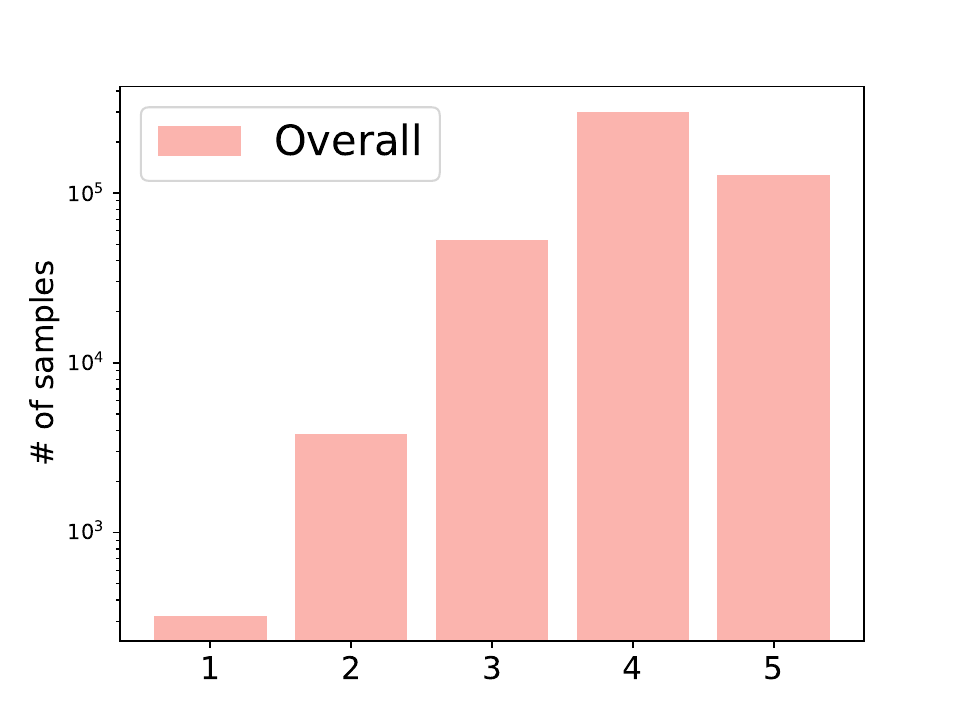}
        \caption{Overall, $\mu=4.14$}
    \end{subfigure}
    \begin{subfigure}[b]{0.32\linewidth}
        \includegraphics[width=\linewidth]{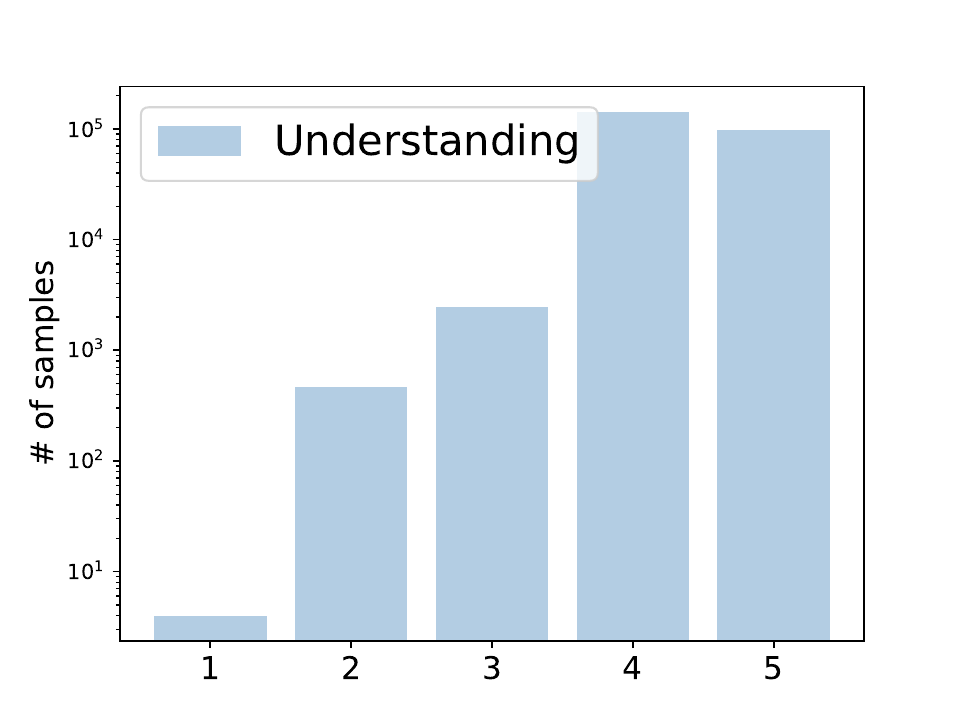}
        \caption{Understanding, $\mu=4.39$}
    \end{subfigure}
    % \hspace{-2mm}
    \begin{subfigure}[b]{0.32\linewidth}
        \includegraphics[width=\linewidth]{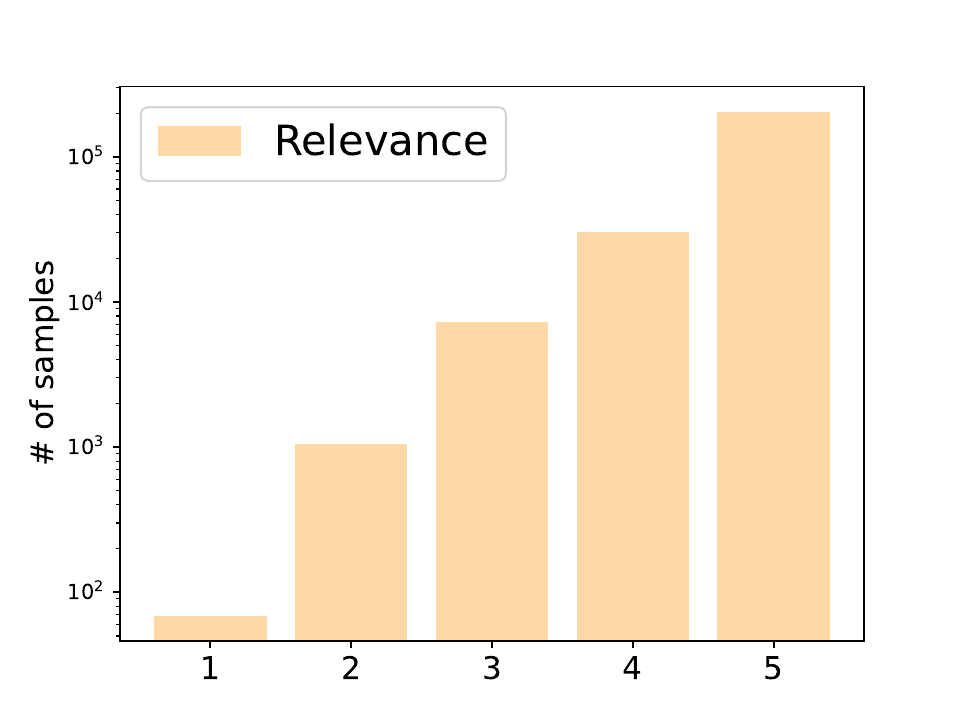}
        \caption{Relevance, $\mu=4.80$}
    \end{subfigure}
    % \hspace{-2mm}
    \centering
    \begin{subfigure}[b]{0.32\linewidth}
        \includegraphics[width=\linewidth]{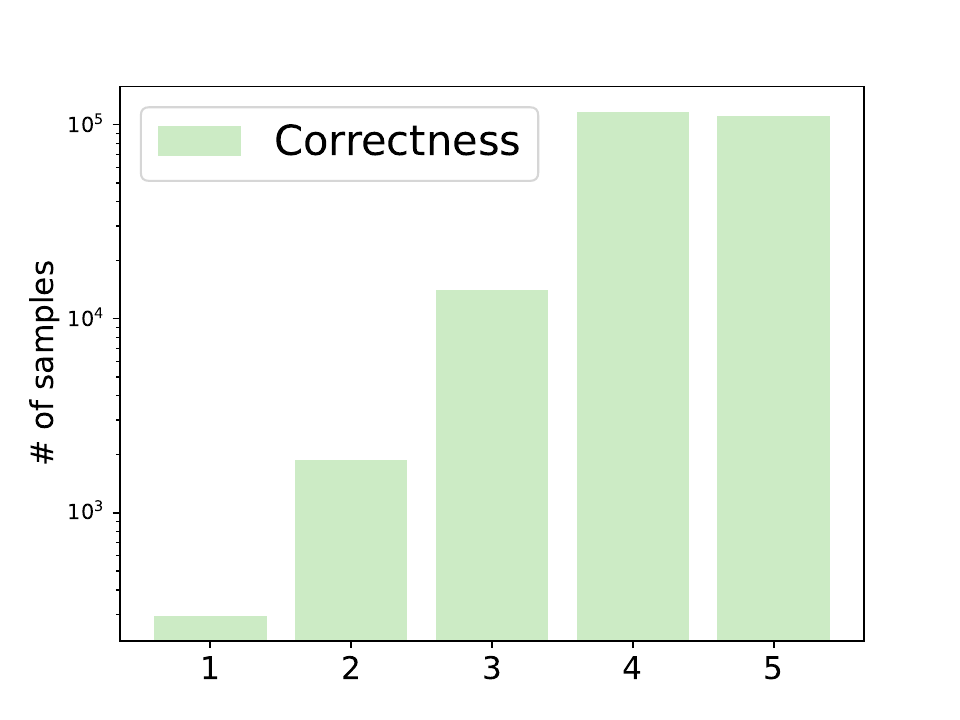}
        \caption{Correctness, $\mu=4.38$}
    \end{subfigure}
    % \hspace{-2mm}
    \begin{subfigure}[b]{0.32\linewidth}
        \includegraphics[width=\linewidth]{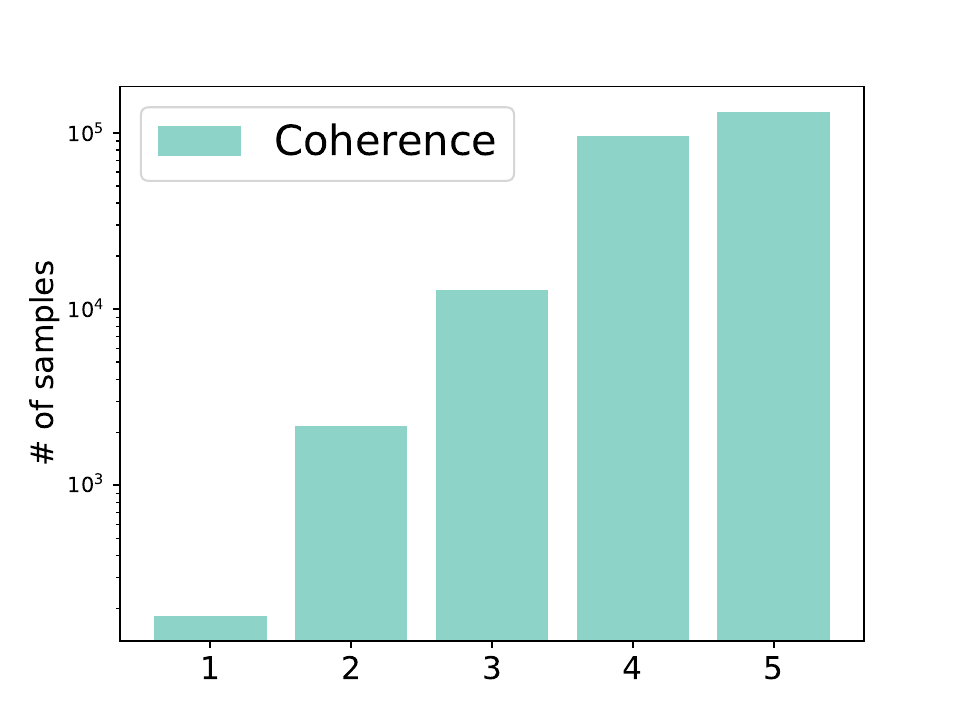}
        \caption{Coherence, $\mu=4.47$}
    \end{subfigure}
    \begin{subfigure}[b]{0.32\linewidth}
        \includegraphics[width=\linewidth]{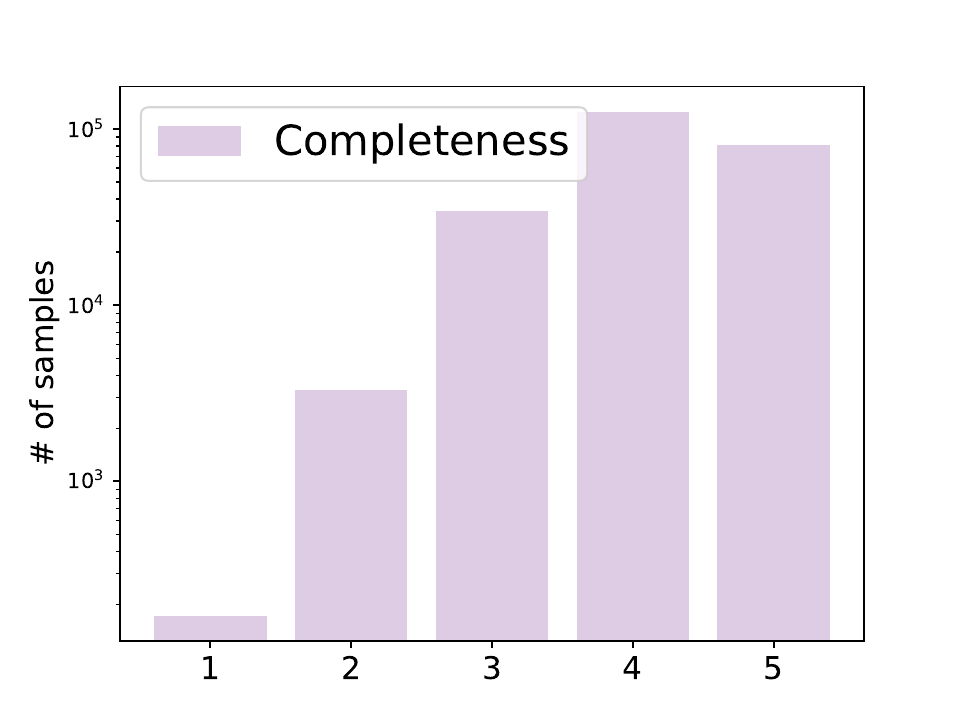}
        \caption{Completeness, $\mu=4.16$}
    \end{subfigure}
    % \hspace{-2mm}
    \caption{The score distribution for the dialogue quality.}
\end{figure*}\label{fig:quality scores}

\subsection{Data Visualization}
% Maybe we can put a table or a figure here to show relevant information.
% \textcolor{blue}{replace datasets table with a figure}

% \textcolor{blue}{Datasets level: turns, average tokens, response tokens}
% \textcolor{blue}{Covered domains: TOD}

The dialogue datasets are compartmentalized into several categories: \textit{Open-Domain Dialogues}, \textit{Task-Oriented Dialogues (TOD)}, \textit{Natural Language Understanding Dialogues (NLU)}, \textit{Conversational Recommendation (Conv-Rec)}, \textit{Dialogue Summarization (Dial-Sum)}, and \textit{Knowledge-Grounded Dialogues (KG-Dial)}. 
% These categorical distinctions are not rigid as some datasets may span multiple categories. For instance, SalesBot encapsulates both casual conversation and task-oriented objectives. Similarly, MultiWOZ, a task-oriented dialogue dataset, incorporates a knowledge base and dialogue acts facilitating knowledge-grounded generation. Furthermore, DialogStudio is marked by its diversity, extending over a wide variety of domains, a selection of which is illustrated in the figure.
%
Figure~\ref{fig:stats_data} presents an overview of DialogStudio's dataset categories.
% they belong to.
% a view of DialogStudio's assortment and domain coverage. 
% We split the dialogue datasets into several categories: \textit{Open-Domain Dialogs}, \textit{Task-Oriented Dialogs (TOD)}, \textit{Natural Language Understanding Dialogs (NLU)}, \textit{Conversational Recommendation Dialogs}, \textit{Knowledge-Grounded Dialogs}. 
Note that the category boundaries are fuzzy as some datasets span multiple categories.
For instance, SalesBot~\citep{chiu2022salesbot} contains both casual and task-oriented conversations. 
Analogously, MultiWOZ~\citep{budzianowski2018large,zang2020multiwoz}, a task-oriented dialogue corpus, incorporates knowledge bases and dialogue acts to enhance knowledge-grounded generation. 
Additionally, DialogStudio demonstrates its diversity by covering a wide range of domains, part of which is shown in Figure~\ref{fig:stats_domain}.  
%
% Moreover, we compare the average length of dialog, user/system utterance for each dataset, in terms of turn number and token number correspondingly
% Figure~\ref{fig:stats_length} shows the . The dialogs in our collection range from less than 3 turns to more than 10 turns, while utterances vary from 3 tokens to nearly 20 tokens.
% Moreover, we compare the average length of dialog, user utterance and system response across 24 datasets as in Figure~\ref{fig:stats_length}.  
% The length of dialogue is calculated based on the number of dialogue turns while the user utterance and system response are computed based on the number of tokens.  

% \begin{figure}
%     \centering
%     \includegraphics[width=0.5\textwidth]{Figs/merged_test.pdf}
%     \caption{Datasets and domain coverages of DialogStudio.}\label{fig:overall}
% \end{figure}

% \begin{figure}
%     \centering
%     \includegraphics[width=1.0\linewidth]{Figs/length.pdf}
%     \caption{Average value of dialogue length, user/system utterance length of each dataset.}\label{fig:stats_length}
% % \vspace{-0.8cm}
% \end{figure}

% \begin{figure}
%     \centering
%     \includegraphics[width=0.5\textwidth]{Figs/domain.png}
%     \caption{Domain coverage of each dataset}
% \end{figure}

\subsection{Data Quality Investigation}
Due to the existence of noise in dialogue, we develop a simple yet effective way to verify the quality of the datasets.
Specifically,
% We select 15-20 TOD datasets, and randomly sample 50 dialogues from each test set.
we employ ChatGPT (GPT-3.5-turbo) to evaluate the quality of system responses based on severall perspectives~\citep{mehri2022report,kim2022soda}, \textit{i.e.}, Understanding, Relevance, Correctness, Coherence, Completeness and Overall quality.
% Informativeness reflects whether the response provides unique and non-generic information that is specific to the dialogue context. 
Understanding assesses whether the model's responses accurately reflect the meaning and intent of the user's inputs. 
Relevance demonstrates whether the generated response should be directly related and appropriate to the preceding user input and the context of the conversation. 
Coherence measures the logical consistency of the model's responses within the context of the conversation.
Completeness refers to whether the system's responses fully address the user's queries or tasks.
% Relevance demonstrates whether the response covers the related domains and intents as in the immediate dialogue.
Overall quality comprehensively rates the quality of dialogue. 
All scores are in the range of 1-5, and higher scores should only be given to truly exceptional examples.
We delicately design the prompt and ask the ChatGPT model to \textit{strictly} rate the score. 

% we view a score greater than 3 as a good quality response. 

Since there are a lot of datasets in DialogStudio, we randomly select 33 multi-turn dialogue datasets and evaluate all the training dialogues of each dataset.
To harmonize ChatGPT and human ratings, we take a random sample of 50 training dialogues from each dataset. These were then rated by three expert researchers using the five specified criteria. Post-alignment of ChatGPT and human evaluations, we view dialogues with a score above 3 as being of high quality.
Figure~\ref{fig:quality scores} illustrates distributions of those scores.
We also reveal the average score as the $\mu$ in each sub-caption.
In general, the dialogues show high qualities regarding to the individual criteria and the overall quality. 

%% file: 4-datasets.tex
\section{Datasets Unification and Access} \label{sec:data}
We collect and process a wide range of datasets, involving different domains, types, and tasks. 
% We show the general unification process in this section and more details are available in GitHub. 
Since these datasets originally contain various information and format, we propose a unification strategy to process all the datasets such that they can be loaded in the same data loader. 

% \subsubsection{format}
\begin{figure*}[h]
\centering
\includegraphics[width=\linewidth]{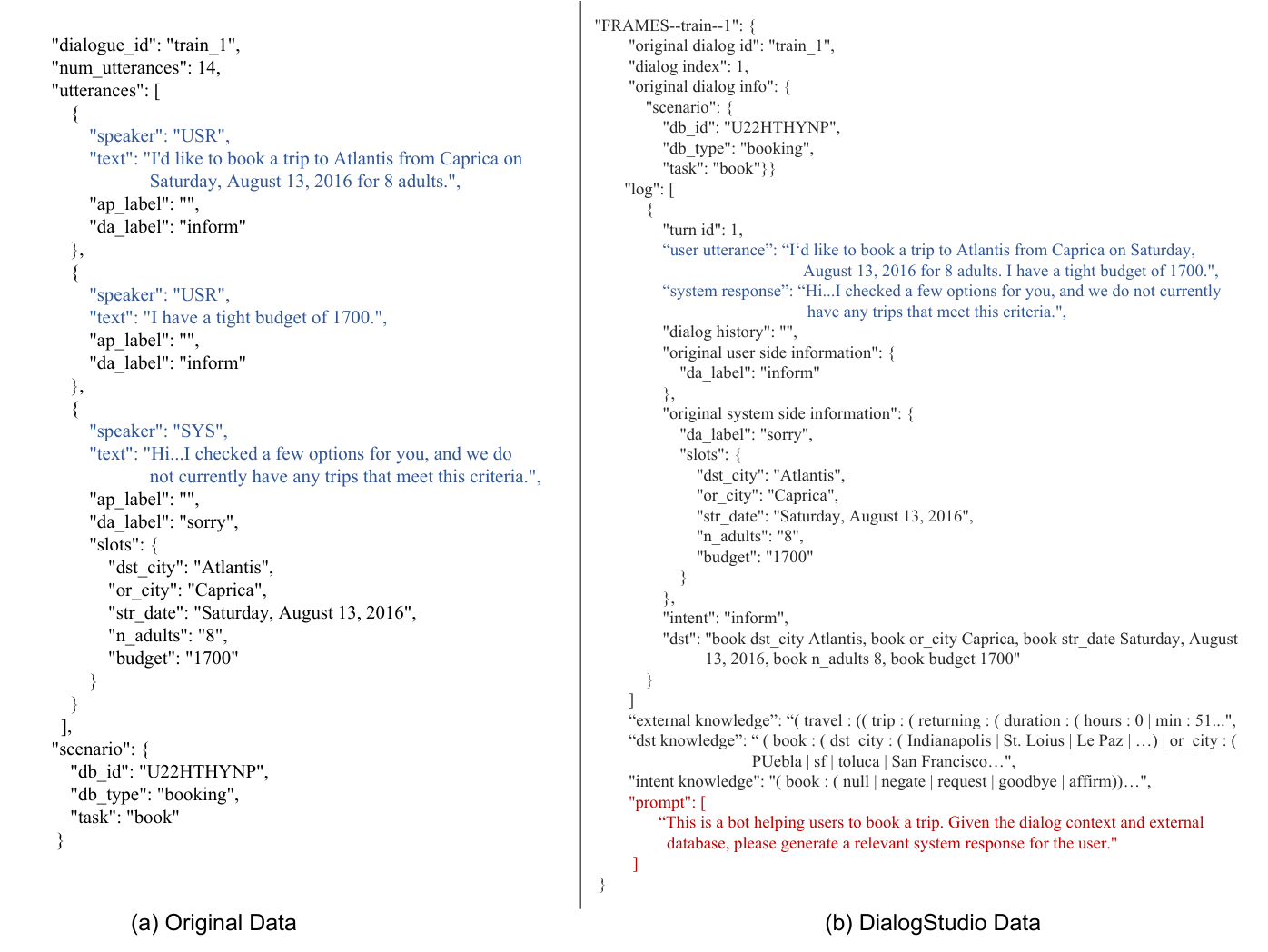}
\caption{A dialogue format example. Left: original example, right: converted example. Here we only show the first turn and partial information.} 
\label{fig:format}
% \vspace{-0.1cm}
\end{figure*}

\subsection{Unification}
% \subsection{Task-Oriented Dialogue Datasets}
% There are total of 16 task-related dialogue datasets in our collection. 
% \noindent\textbf{Split}
% \subsubsection{Split}
% We conduct the train/val/test split based on their original split strategy. If there is no pre-defined split strategy, we consider the whole dataset as the training set.

% \subsubsection{Corrections}
% We fixed dialogs with following issues (less than 1\% dialogs) : (1) We remove dialogs which belong to multi-turn dialogs, but there are only one turn, and only user utterances (i.e., lacks of system responses) or only system responses. (2) Some dialogs contains one or more empty user or system utterances during a conversation. We manually check the individual dialogs and fill with utterances based on corresponding dialogue contexts, dialogue acts, dialogue information, etc.
Before unifying the format of those datasets, we fixed several issues as follows: 1) we remove those dialogues labeled as multi-turn dialogues, but actually with only one turn and miss either user utterance or system utterance. 2) We manually check the individual dialogues. If one dialogue contains one or more empty user or system utterances, we fill utterances based on corresponding dialogue contexts, dialogue acts, and dialogue information. In total, less than $0.5\%$ of dialogues had these issues. To support research interest on individual datasets, we have flagged and rectified these problematic dialogues.

Additionally, we recognize the success of instruction tuning for dialogue models and thus we manually pre-define five different prompt templates for multi-turn dialogue datasets, such as \textit{This is a bot helping users to \{Task\_Domain\}. Given the dialogue context and external database, please generate a relevant system response for the user.} The \textit{\{Task\_Domain\}} is associated with the dialogue domain and we manually create a corresponding description. 
For example, if a dialogue is of domain \textit{travel}, we set \textit{\{Task\_Domain\}} as \textit{book a trip}. 
A concrete example of the prompt is demonstrated in Figure~\ref{fig:format}. Moreover, many datasets lack a direct mapping between dialogues and their domain information. To address this, we determine the domain of each dialogue using its intent, schema, APIs, and associated databases.

% Therefore, for each dialog, we sample five prompt template and fill in the combination of domain descriptions, corresponding to its involved domains. For single-turn dialogue datasets such as NLU datasets, we provide one prompt for each task. Fig.~\ref{fig:format} shows one example under the "prompt". 

% \subsubsection{Prompt}
% For multi-turn dialogue datasets, we manually pre-define five different prompt templates, such as "This is a bot helping users to $\_\_\_$ . Given the dialogue context and external database, please generate a relevant system response for the user." For each domain, we also write a corresponding description, e.g. "restaurants": "find and book a restaurant". Therefore, for each dialog, we sample five prompt template and fill in the combination of domain descriptions, corresponding to its involved domains. For single-turn dialogue datasets such as NLU datasets, we provide one prompt for each task. Fig.~\ref{fig:format} shows one example under the "prompt". 

Next, we construct a uniform JSON dictionary format to store all relevant information of each dialogue as illustrated in Figure~\ref{fig:format}. 
% \footnote{More examples are available in the supplementary data materials.}  
% We believe this format is a more generalized version 
Compared with existing works,
DialogStudio covers more dialogue information and is easier to retrieve the information for arbitrary dialogue-related tasks.
Concretely, we include all dialogue-related information, such as the dialogue ID, data split label, domain, task, and content.
Additionally, we identify the external knowledge, dialogue state tracking (DST) knowledge, and intent knowledge in the dialogue, which are the most beneficial knowledge for a dialogue. 

Regarding external knowledge, we construct it based on information such as databases and dialogue acts.
Since each dialogue dataset focuses on specific tasks or domains and has a different database and annotation schema, we unify  such information into \textit{external knowledge}. 
For example, if the user is looking for a hotel and asking for its address, the system response should be based on both the search results from the database and the dialogue context. 
To simulate the realistic situation and avoid directly providing the model with the ground truth resulting hotel, we also randomly sample four other candidate results and mix them with the ground truth result.
All information is flattened and converted into a string as external knowledge.  

To complete tasks and generate coherent responses, a dialogue system needs to track users' requirements for the task. 
Those requirements are usually represented as dialogue states.
For example, regarding the hotel booking task, a dialogue system needs to extract information such as price range and locations to enable searching hotels in the database. 
The type of dialogue states varies across different tasks and datasets. 
As such, it is hard for dialogue systems to predict the values of those dialogue states if unknowing the specific dialogue states the task covers. 
Therefore, we propose to insert the schema, consisting of pre-defined dialogue state types and values for each task, into the input sequence.
For datasets like SGD~\citep{rastogi2020towards}, which already provides annotation schema, we directly convert the dictionary-structured schema into a string. 
For the rest datasets that have no such schema file, we iterate over all dialogues and collect potential state annotations to construct a schema.
We provide domains, slot types, and slot values in the schema string. 
For those categorized dialogue slots like "hotel star-ratings", which have a fixed number of candidate values, we provide all possible values.
For others that have unlimited possible values, e.g. "stay night", we randomly sample ten values, such that a model can learn what slot values are relevant to these slot types. 
We put the turn-level ground-truth DST information in "dst", and the general DST information under "dst knowledge", as presented in Figure~\ref{fig:format}.

Analogously, intent prediction also requires models to know all possible intent types for each task. 
Therefore, we extract the schema directly from the schema file if it exists. As to datasets without schema, we also iterate over all dialogue in the dataset to collect all potential intents.
Then, we put the turn-level ground-truth intent information into "intent", and the general intents under "intent knowledge", as presented in Figure~\ref{fig:format}.
Note that not all datasets provide detailed annotation for dialogue states, intents, or even databases. 
For dialogue state tracking and intent classification tasks, we only process dialogues with corresponding annotations. 
Since all data is used for response generation, we leave the external knowledge value for the database blank if there is no related database in the original dataset.

\subsection{Access and Maintenance}
% \textcolor{red}{Formats are great; Each has 5 examples; Easy to load on HuggingFace; document for each dataset; all code for pre-processing, cleaning are public}
% \subsubsection{Utility and Availability}
As aforementioned in the format, our DialogStudio data is easy to access via the JSON files.
To make DialogStudio more maintainable and accessible, we will publish datasets on both GitHub and HuggingFace.
GitHub mainly stores selected dialogue examples and relevant documents. We sample five original dialogues and five converted dialogues for each dataset to facilitate users in comprehending our format and examining the contents of each dataset.
The complete DialogStudio dataset is maintained in our HugginFace repository, where all the datasets can be directly downloaded or loaded with the HuggingFace $\mathtt{load\_dataset(dialogstudio,dataset\_name)}$ API. Given the substantial volume of datasets, optimizing user experience poses a challenge and limitation.  We will continuously maintain and update both GitHub and HuggingFace.
% \begin{align*}
% \mathtt{load\_dataset(``Salesforce/dialogstudio",} \\
% \mathtt{``dataset\ name")}
% \end{align*}
DialogStudio is built upon public research datasets without individual or private information. We believe it is important to clearly present the license associated with each of these datasets. Consequently, we have included the original licenses for all datasets. All these datasets are supportive of academic research, and some of them also endorse commercial usage. The code that we employ falls under the widely accepted Apache 2.0 license.  While we strictly require adherence to the respective dataset licenses for all intended usages on DialogStudio, there remains a possibility that some  works might not fully comply with the licenses. 

Regarding the other concerns such as ethical concern, we admit that DialogStudio is collected and maintained by the authors of this work and we did not hire external annotators. 
Since it contains unified datasets across several categories, it supports various research purposes from individual tasks and datasets to language model pre-training.  

%% file: 6-exp_results.tex
\section{Experiments} \label{sec:eval}
In this section, we  present the pre-training details, methodologies, and metrics used to assess the performance of our DialogStudio model. The evaluation process aims to measure the model's ability to both solve task-oriented dialogues and understand general prompt-based instruction. 

\input{5-exp_settings}

\begin{table*}[]
\centering
% \small
% \resizebox{0.8\linewidth}{!}{
% \scalebox{1.0}{
\begin{tabular}{l|cc|cc}
\toprule
\hline
% \hline
                    & \multicolumn{2}{c|}{CoQA} & \multicolumn{2}{c}{MultiWOZ} \\ \cline{2-5}
                    & ROUGE-L         & F1           & ROUGE-L         & F1          \\ \hline
Flan-T5-3B~\citep{longpre2023flan}         &    37.1           &   37.2        &   7.0         & 7.4     \\
Flan-T5-Large~\citep{longpre2023flan}         &    22.5           &   22.3        & 15.9             & 17.6          \\
GODEL-Large~\citep{peng2022godel}          & 43.2          & 43.3      & 18.5             & 19.3          \\ \hdashline
DialogStudio-T5-Large  & 61.2          & 61.5      & 32.4             & 34.5          \\     
DialogStudio-Flan-T5-Large  & 63.3          & 63.5      & 33.7             & 35.9          \\ 
\bottomrule
\end{tabular}\caption{Zero-shot results on CoQA and MultiWOZ 2.2.}\label{table:multiwoz}

% \vspace{-0.2cm}
\end{table*}

\begin{table*}[]
\resizebox{1.0\linewidth}{!}{
\centering
\begin{tabular}{l|ccc|cc}
\toprule
\hline
                 & \begin{tabular}[c]{@{}c@{}}CR \\ (14 tasks)\end{tabular} & \begin{tabular}[c]{@{}c@{}}DAR \\ (7 tasks)\end{tabular} & \begin{tabular}[c]{@{}c@{}}TE \\ (27 tasks)\end{tabular} & \begin{tabular}[c]{@{}c@{}}avg. \\ (48 tasks) \end{tabular}    \\ \hline
OPT-30B~\citep{zhang2022opt}          & 21.3/1.1                        & 35.2/4.1                        & 40.3/0.9                    & 32.3/2.0                                               \\
OPT-IML-30B~\citep{iyer2022opt}      & 37.4/41.6                        & 51.4/51.8                        & 54.7/47.8                 & 47.9/47.1                                          \\
OPT-175B~\citep{zhang2022opt}         & 21.0/4.2                      & 37.1/16.8                  & 41.6/2.2                & 33.3/7.7                                        \\
OPT-IML-175B~\citep{iyer2022opt}     & 39.0/49.8                     & \textbf{61.2}/60.2                        & 54.3/51.0                    & \textbf{51.5}/53.6                                       \\ 
Tk-INSTRUCT-11B~\citep{wang2022super}  & 32.3/\textbf{62.3}                  & 51.1/\textbf{69.6}                        & \textbf{55.0}/\textbf{64.1}              & 46.1/\textbf{65.3}                                       \\ \hdashline
Tk-INSTRUCT-3B~\citep{wang2022super}   & 38.4/51.3                     & 45.7/58.5                           & 48.4/52.8                   & 44.2/54.2                          \\ 
DialogStudio-NIV2-T5-3B & \textbf{41.3}/59.8                       & 57.5/63.7                        & 52.3/55.1              & 50.4/59.5                     \\ 
\bottomrule
\end{tabular}}
\caption{0-shot/2-shot/5-shot ROUGE-L testing results on unseen datasets and unseen tasks. Results of baselines are reported by original papers. CR, DAR, and TE, avg. are abbreviations for Coreference Resolution,
Dialogue Act Recognition, 
Textual Entailment, and average results, respectively.}\label{table:niv2}
% \vspace{-0.2cm}
\end{table*}

\subsection{Evaluation for Response Generation}
% \noindent\textbf{Evaluation for Response Generation}

% GODEL~\citep{peng2022godel} provides a benchmark to evaluate model's ability for task-oriented dialogs and knowledge grounding.

% \paragraph{COQA} 
% COQA~\citep{reddy2019coqa} is a multi-turn conversational question answering dataset that contains 127,000 questions with answers collected from more than 8000 conversations. Each conversation is collected by pairing two crowdworkers to chat about a passage in the form of conversational questions and answers. 

% \textbf{Wizard of Wikipedia}
% is a dataset of multi-turn knowledge grounded di- alogs between an apprentice and a wizard, who has access to Wikipedia sentences and labels the sentences utilized for each utterance they made. To focus on grounded response generation, we use the gold Wikipedia sentences provided at each turn of the dialog.

% \textbf{Wizard of Internet}
% is an open-domain dialogue dataset grounded on internet retrieved sentences. At each turn, the wizard can issue a free text web search and replies to the ap- prentice grounding the response on the retrieved sentences. Similarly, we use the gold retrieved sentences provided at each turn of the dialog.
\noindent\textbf{Settings.}
We evaluate the performance on CoQA~\citep{reddy2019coqa} and MultiWOZ 2.2~\citep{zang2020multiwoz}. CoQA is a multi-turn conversational question answering dataset  with 8k conversations about text passages from seven diverse domains. MultiWOZ 2.2 is one of the largest  and  most widely used multi-domain task-oriented dialogue corpora with more than 10000 dialogues. Each dialogue involves with one or more domains such as \textit{Train, Restaurant, Hotel, Taxi,} and \textit{Attraction}. The dataset is challenging and complex due to the multi-domain setting and diverse linguistic styles.  Note that we exclude both datasets during the pre-training stage to prevent data leakage.

For CoQA, we follow the original paper setting to answer question based on external passage. For MultiWOZ 2.2,
we consider the lexicalized dialogue-act-to-response generation task where the model needs to consider both the dialogue context and the dialogue acts during generation. We follow the prompt from \citep{bang2023multitask} to instruct models, i.e., \textit{Continue the dialogue as a task-oriented dialogue system called SYSTEM. The answer of SYSTEM should follow the ACTION provided next while answering the USER’s last utterance}.

We focus on zero-shot  evaluation and report the ROUGE-L and F1 score~\citep{miller2017parlai}, where ROUGE-L measures the  longest common subsequence and F1 measures the Unigram F1 overlap between the prediction and ground-truth response. 
% For the few-shot setting, we sample 50 dialogues from the training set~\cite{peng2022godel} to fine-tune the model and use the original test set for evaluation. 

\begin{table*}[]
\centering
% \small
% \resizebox{0.8\linewidth}{!}{
% \scalebox{1.0}{
\begin{tabular}{l|cc|cc}
\toprule
\hline
% \hline
                    & \multicolumn{2}{c|}{MMLU} & \multicolumn{2}{c}{BBH} \\ \cline{2-5}
                    & 0-SHOT        & 5-SHOT   & 3-SHOT        \\ \hline
% LLAMA 7B~\citep{touvron2023llama}  &        & 35.2	     &        30.9         \\
TK-INSTRUCT 11B~\citep{wang2022super} &- & 41.1 & 32.9 \\
LLAMA 13B~\citep{touvron2023llama}  &    -    & 46.2	     &        37.1           \\ 	
Vicuna 13B~\citep{chiang2023vicuna}  &    -    & 49.7	     &        37.1    \\
Flan-T5-Large~\citep{longpre2023flan}        &    41.5           &       41.9    &         37.1            \\
Flan-T5-XL~\citep{peng2022godel}          &     48.7    & 49.3      &        40.2            \\
OPT-IML-Max 30B~\citep{iyer2022opt}  &     46.3    & 43.2      &        31.3           \\ \hdashline
DialogStudio-Flan-T5-Large  &   40.1       &  40.9  &     34.2              \\ 
DialogStudio-Flan-T5-3B  &   48.3       &  47.8    & 40.3                  \\     
\bottomrule
\end{tabular}\caption{Test results on MMLU and BBH. Results come from original papers and InstructEval~\citep{chia2023instructeval}.}\label{table:mmlu}
\end{table*}

\noindent\textbf{Baselines.}
We consider GODEL~\citep{peng2022godel} and Flan-T5~\citep{longpre2023flan} as our baselines. GODEL is a T5-based  large pre-trained model for goal-oriented dialogues. It is pre-trained with 551M multi-turn Reddit dialogues and 5M knowledge-grounded and question-answering dialogues. Flan-T5 is an instruction-aware model. It is also initialized from T5 and pre-trained on the Flan collection, which consists of more than 1800 tasks and 400 datasets, including dialogue datasets. 

\noindent\textbf{Results.}
Table \ref{table:multiwoz} depicts the results from both zero-shot and few-shot testing. Evidently, our models considerably surpass the baseline models in terms of zero-shot learning, exhibiting a robust generalized ability for response generation in a zero-shot scenario. 

Flan-T5-3B, on the other hand, underperforms in the task of generating responses from dialog-acts. This model tends to produce incorrect dialog acts, unnatural utterances, or terminates with an empty end token. One  explanation for this is that Flan-T5 models did not receive sufficient dialogue training during the instruction-training phase on the Flan collections. Comparisons between the performances of existing models before and after training on the unified dataset validate DialogStudio's usefulness. 

\subsection{Evaluation on Super-NaturalInstructions
}

\noindent\textbf{Settings.} NIV2~\citep{wang2022super} introduces an instruction-tuning benchmark with more than 1600 tasks. We select 3 categories with 44 tasks from the held-out test set, which consists of 154 tasks, i.e., Coreference Resolution, Dialogue Act Recognition, and Textual Entailment.  The selected tasks and datasets are unseen in the training stage.
Specifically, we strictly follow all settings including metrics in \cite{wang2022super}, i.e., train models with instructions + 2 positive demonstrations and no negative demonstrations. We fine-tune DialogStudio-T5-3B on 756 training tasks.

\noindent\textbf{Baselines.}
OPT~\citep{zhang2022opt} is a set of open  decoder-only transformer models pre-trained on 180B tokens. 
OPT-IML~\citep{iyer2022opt} is an instruction meta-learning model based on the OPT-IML bench with more than 1500 tasks.
Tk-INSTRUCT~\citep{wang2022super} is initialized from T5 and further pre-trained based on NIV2 collections.
Note that we neglect Flan-T5 because it trains with all the downstream datasets and tasks. 

\noindent\textbf{Results.}
Table \ref{table:niv2} shows the 0-shot and 2-shot results on unseen datasets and unseen tasks. Based on the average performance on 48 tasks, DialogStudio-NIV2-T5-3B outperforms OPT-IML-175B by 5.9\% on 2-shot learning with more than 50 times fewer model parameters, and it surpasses Tk-INSTRUCT-11B by 4.3\% on 0-shot learning with more than 3 times fewer parameters. The performance demonstrates the strong generalization ability of DialogStudio model. 
Compared with Tk-INSTRUCT-3B, DialogStudio-NIV2-T5-3B achieves 6.2\% and 5.3\% improvements on 0-shot and 2-shot learning respectively.
Given that both Tk-INSTRUCT and our  DialogStudio-NIV2-T5-3B  are fine-tuned from the T5 model, this improvement indicates the effectiveness of pre-training with our DialogStudio collection. 

% \textcolor{red}{Will put a table here to compare and analyze with OPT-IML, GODEL, DialogOhana}

\subsection{Evaluation on MMLU and BBH}

Table \ref{table:mmlu} presents results on MMLU~\citep{hendrycks2020measuring} and Big Bench Hard 
 (BBH)~\citep{srivastava2022beyond}. When comparing the performance of DialogStudio-Flan-T5 with Flan-T5, we observe a minor decrease. However, this is accompanied by a significant improvement in dialogue relevant capabilities.

\subsection{Evaluation on Alternative Benchmarks}
DialogStudio encompasses not just public realistic dialogue datasets, but also those derived from or shared with ChatGPT, such as SODA~\cite{kim2022soda} and ShareGPT. Due to GPU constraints, we employ techniques like LoRA~\cite{hu2021lora} to fine-tune llama~\cite{touvron2023llama}. When using equivalent datasets from DialogStudio, we observed performance comparable to other models, e.g., Vicuna~\cite{chiang2023vicuna}, on benchmarks like AlpacaEval~\cite{dubois2023alpacafarm} and MT-Bench~\cite{zheng2023judging}. This demonstrates that DialogStudio caters to research interests in both specific datasets and generalized instruction tuning. 

% \subsection{Evaluation for General Understanding}
% % MMLU

% \subsection{Evaluation for General Generation}
% % 

% Datasets
% GODEL: Four datasets (CoQA, WOW, WOI, MultiWOZ)
% In-context MultiWOZ DST
% OPT-IML: 44 Tasks (coreference resolution, dialogue act recognition, text entailment)
% Intent detection (Add later)

%MMLU 

%% file: 5-exp_settings.tex
\subsection{Model Pre-training}

%MS-DC~\citep{li2018microsoft},
%  PLACES3.5~\citep{}
In this section, we introduce more details about how we conduct our pre-training. 
In regards of training models, we  mix several datasets from DialogStudio.

For task-oriented and conversational recommendation datasets, we selected dialogues from a range of sources including KVRET~\citep{eric2017key}, AirDialogue~\citep{wei2018airdialogue}, DSTC2-Clean~\citep{mrkvsic2017neural}, CaSiNo~\citep{chawla2021casino}, FRAMES~\citep{elframes}, WOZ2.0~\citep{mrkvsic2017neural}, CraigslistBargains \citep{he2018decoupling}, Taskmaster1-2~\citep{byrne2019taskmaster}, ABCD~\citep{chen2021action}, MulDoGO~\citep{peskov2019multi}, BiTOD \citep{lin2021bitod},   SimJoint~\citep{shah2018building},  STAR \citep{mosig2020star}, SGD~\citep{rastogi2020towards}, OpenDialKG~\citep{moon2019opendialkg} and DuRecDial-2.0 \citep{liu2021durecdial}.

Meanwhile, for knowledge-grounded dialogues, we drew upon dataset from SQA~\citep{iyyer2017search}, SParC~\citep{yu2019sparc}, FeTaQA~\citep{nan2022fetaqa}, MultiModalQA~\citep{talmor2021multimodalqa}, CompWebQ~\citep{talmor2018web}, CoSQL~\citep{yu2019cosql}.

For open-domain dialogues, we sample dialogues from SODA~\citep{kim2022soda}, ProsocialDialog~\citep{kim2022prosocialdialog}, Chitchat~\citep{myers2020conversational}. 

For each dialogue dataset, we sample at most 11k dialogues. Additionaly, we extracted around 11k dialogue turns from question-answering dialogues featured in RACE~\citep{lai2017race}, NarrativeQA~\citep{kovcisky2018narrativeqa}, SQUAD~\citep{rajpurkar2018know}, MCtest~\citep{richardson2013mctest}, OpenBookQA~\citep{OpenBookQA2018}, MultiRC~\citep{MultiRC2018}. Here, a dialogue turn refers to a pair consisting of a dialogue context and its corresponding system response. 
The rest datasets in DialogStudio are preserved for future evaluations and downstream fine-tuning. 

For each dialogue during the training, we shape the available external knowledge into a string, which is included in dialogue context, and instruct the model to generate a dialogue response based on external knowledge. We use the format \textit{Instruction \textbackslash{}n \textless{}USER\textgreater{} user utterance \textless{}SYSTEM\textgreater{} system response \textless{}USER\textgreater{} ...  \textless{}USER\textgreater{} user utterance \textbackslash{}n \textless{}EXTERNAL KNOWLEDGE\textgreater{} supported knowledge} to train the model, where \textit{\textless{}USER\textgreater{}}, \textit{\textless{}SYSTEM\textgreater{}} and \textit{\textless{}EXTERNAL KNOWLEDGE\textgreater{}} are special tokens. 

% To boost model's robustness and adaptability to different input structures, we also sample $20\%$ of the used ToD data and change the relevant order of \textit{("Prompt", "External Knowledge", "Dialogue Context")}. This introduces more variety in the input format. The dialogue pre-training comprises approximately 6 million dialogue turns.

We follow the public HuggingFace transformer code~\citep{wolf2020transformers,wang2022super} to train the model. 
For initializing our models, we adopt T5~\citep{raffel2020exploring} and Flan-T5~\citep{longpre2023flan} as starting points to respectively establish DialogStudio-T5 and DialogStudio-Flan-T5. For the training of DialogStudio-Flan-T5, we exclude all translation-oriented tasks, limiting the sample size for each Flan task to a maximum of 150 examples. This leads to a cumulative total of 140,000 samples. We train the model up to 3 epochs with bfloat16 precision, a total batch size of 64. We set a constant learning rate 5e-5  and 3e-5 for the large model and the 3B model, respectively.  Experiments are conducted using 16 A100 GPUs, each with 40GB of GPU memory.

%% file: 7-conclusion.tex
\section{CONCLUSION}
% In this paper, we introduce DialogStudio, a unified and diverse collection of 80 dialogue datasets, retaining their original information.
% Utilizing DialogStudio, we developed corresponding models, demonstrating superior performance in both zero-shot and few-shot learning scenarios. 
% We will release DialogStudio  to foster transparency and support research into datasets, tasks, and pre-training models for conversational AI.

In this study, we have introduced DialogStudio, a comprehensive collection that aggregates more than 80 diverse dialogue datasets while preserving their original information. This aggregation not only represents a significant leap towards consolidating dialogues from varied sources but also offers a rich tapestry of conversational patterns, intents, and structures, capturing the nuances and richness of human interaction. Utilizing DialogStudio, we developed corresponding models, demonstrating superior performance in both zero-shot and few-shot learning scenarios. In the spirit of open research and advancing the field, we are committed to releasing DialogStudio to the broader research community. 
% We believe the release will catalyze further innovations, facilitating transparency, and propelling research into not just the datasets but also diverse tasks and the development of robust pre-training models in the conversational AI domain. 
% Through this initiative, we wish to build a solid foundation and set new benchmarks for future endeavors in this ever-evolving field.

%% file: 8-appendix.tex
Table \ref{dialogstudio:table} and Table \ref{dialogstudio:table_2} lists datasets included in DialogStudio. Initially, we present a partial list of these datasets. More and latest information are available in GitHub\footnote{\url{https://github.com/salesforce/DialogStudio}}.

% \usepackage{multirow}
\begin{table*}[ht]
\begin{tabular}{c|l}
\hline
\multirow{17}{*}{\textbf{NLU}} & NLU++~\citep{casanueva2022nlu++} \\ 
                               & BANKING77-OOS~\citep{zhang2022pre}                                          \\ 
                               & BANKING77~\citep{casanueva2020efficient}                                              \\
                               & RESTAURANTS8K~\citep{coope2020span}                          \\
                               & CLINC150~\citep{larson2019evaluation}                                               \\
                               & CLINC-Single-Domain-OOS-banking~\citep{zhang2022pre}                        \\  
                               & CLINC-Single-Domain-OOS-credit\_cards~\citep{zhang2022pre}                 \\
                               & HWU64~\citep{liu2019benchmarking}                                                  \\
                               & SNIPS~\citep{coucke2018snips}                                                  \\
                               & SNIPS-NER~\citep{coucke2018snips}                                               \\
                               & DSTC8-SGD~\citep{coope2020span}                              \\
                               & TOP~\citep{gupta2018semantic}                                           \\
                               & TOP-NER~\citep{gupta2018semantic}                                                \\
                               & ATIS-NER~\citep{hemphill1990atis}                                               \\
                               & ATIS~\citep{hemphill1990atis}                                                   \\
                               & MIT-MOVIE~\citep{liu2013asgard}                                              \\
                               & MIT-RESTAURANT~\citep{liu2013asgard}   \\ \hline
\multirow{27}{*}{\textbf{TOD}} & KVRET~\citep{eric2017key} \\ 
                               & AirDialogue~\citep{wei2018airdialogue}                                          \\ 
                               & DSTC2-Clean~\citep{mrkvsic2017neural}                                          \\
                               & CaSiNo~\citep{chawla2021casino}                       \\
                               & FRAMES~\citep{elframes}                        \\  
                               & WOZ2.0~\citep{mrkvsic2017neural}                \\
                               & CraigslistBargains \citep{he2018decoupling}                                                \\
                               & Taskmaster1~\citep{byrne2019taskmaster}                                                 \\
                               & Taskmaster2~\citep{byrne2019taskmaster}                                               \\
                               & Taskmaster3~\citep{byrne2019taskmaster}                              \\
                               & ABCD~\citep{chen2021action}                                         \\
                               & MulDoGO~\citep{peskov2019multi}                                              \\
                               & BiTOD \citep{lin2021bitod}                                              \\
                               & SimJointGEN~\citep{shah2018building}                                                  \\
                               & SimJointMovie~\citep{shah2018building}                                               \\
                               & SimJointRestaurant~\citep{shah2018building}  \\
                               & STAR \citep{mosig2020star} \\
                               & SGD~\citep{rastogi2020towards} \\
                               & MultiWOZ2\_1~\citep{eric2020multiwoz} \\ 
                               & MultiWOZ2\_2~\citep{zang2020multiwoz} \\
                               & MultiWOZ2\_2+~\citep{Qian2021AnnotationIA}\\
                               & HDSA-Dialog~\citep{chen2021action} \\
                               & MS-DC~\citep{li2018microsoft} \\
                               & GECOR~\citep{quan2019gecor} \\
                               & Disambiguation~\citep{qian2022database} \\
                               & MetaLWOZ~\citep{lee2019multi} \\
                               & KETOD~\citep{chen2022ketod} \\
                               & MuDoCo~\citep{martin2020mudoco} \\ \hline

\end{tabular}\caption{List of datasets included in DialogStudio (a).}\label{dialogstudio:table}
\end{table*}

% %%%%%%%%%%%%%%%%%%%%%%%%%%%%%%%
\begin{table*}[ht]
\begin{tabular}{c|l}
\hline
\multirow{19}{*}{\textbf{KG-Dial}}   &       
     SQA~\citep{iyyer2017search} \\              
     & SParC~\citep{yu2019sparc} \\
     & FeTaQA~\citep{nan2022fetaqa} \\
     & MultiModalQA~\citep{talmor2021multimodalqa} \\
     & CompWebQ~\citep{talmor2018web} \\
     & CoSQL~\citep{yu2019cosql} \\ 
     & CoQA~\citep{reddy2019coqa} \\
     & Spider~\citep{yu2018spider} \\
     & ToTTo~\citep{parikh2020totto} \\
     & WebQSP~\citep{yih2016value} \\
     & WikiSQL~\citep{zhong2017seq2sql} \\
     &WikiTQ~\citep{pasupat2015compositional} \\
     & DART~\citep{nan2021dart} \\
    & GrailQA~\citep{gu2021beyond} \\
     & HybridQA~\citep{chen2020hybridqa} \\
     & MTOP~\citep{chen2020hybridqa} \\
     & UltralChat-Assistance~\citep{ding2023enhancing} \\ 
    & Wizard\_of\_Wikipedia~\citep{dinan2018wizard} \\
     & Wizard\_of\_Internet~\citep{komeili2022internet} \\ \hline
\multirow{12}{*}{\textbf{Dial-Sum}} & TweetSumm~\citep{feigenblat-etal-2021-tweetsumm-dialog}\\
                    & SAMSum~\citep{gliwa-etal-2019-samsum}                                              \\
                    & DialogSum~\citep{chen-etal-2021-dialogsum}                                           \\
                    & AMI~\citep{kraaij2005ami, rennard2023abstractive}                                                 \\
                    & ICSI~\citep{janin2003icsi}                                                \\
                    & QMSum~\citep{zhong2021qmsum}                                               \\
                    & MediaSum~\citep{zhu2021mediasum}                                            \\
                    & ECTSum~\citep{mukherjee-etal-2022-ectsum}                                              \\
                    & SummScreen\_ForeverDreaming~\citep{chen-etal-2022-summscreen}                         \\
                    & SummScreen\_TVMegaSite~\citep{chen-etal-2022-summscreen}                              \\
                    & CRD3~\citep{rameshkumar-bailey-2020-storytelling}                                                \\
                    & ConvoSumm~\citep{fabbri-etal-2021-convosumm}                                           \\ \hline
\multirow{8}{*}{\textbf{Open-Domain}}   & ChitCHAT~\citep{myers2020conversational} \\
                                  & SODA~\cite{kim2022soda} \\
                                  & Prosocial~\citep{kim2022prosocialdialog}\\
                                  & HH-RLHF~\citep{bai2022training}\\
                                  & Empathetic~\citep{rashkin2019towards}\\
                                  & ConvAI2~\citep{Dinan2019TheSC}\\
                                  & AntiScam ~\citep{li2020end}\\
                                  & ShareGPT ~\citep{zheng2023judging} \\
                                  & PLACES3.5~\citep{chen2023places}\\ \hline
  \multirow{5}{*}{\textbf{Conv-Rec}} & SalesBot~\citep{chiu2022salesbot} \\    
                                  & Redial~\citep{li2018conversational}   \\
                                  & Inspired~\citep{hayati2020inspired} \\
                                  & DuRecDial 2.0~\citep{liu2021durecdial}   \\
                                  & OpendialKG~\citep{moon2019opendialkg}     \\ \hline
\end{tabular}\caption{List of datasets included in DialogStudio (b).}\label{dialogstudio:table_2}
\end{table*}

%% file: supplementary.tex
\appendix

%\section*{Appendix}

\section{Additional Notes on the Use of NLI}
\label{app:nli}

There are other tasks modeling relationships between sentences.
Paraphrase~\citep{paranmt} and semantic relatedness~\citep{semeval} tasks are such examples.
It is possible to automatically create large-scale paraphrase datasets by machine translation~\citep{ppdb}.
However, our task is not a paraphrasing task, and creating negative examples is crucial and non-trivial~\citep{selectional-preference}.
In contrast, as described above, the NLI setting comes with negative examples by nature.
The semantic relatedness (or textual similarity) task is considered as a coarse-grained task compared to NLI, as discussed in the previous work~\citep{jmt}, in that the task measures semantic or topical relatedness.
This is not ideal for the intent detection task, because we need to discriminate between topically similar utterances of different intents.
In summary, the NLI task well matches our objective, with access to large datasets.

\section{A Note on the Threshold Selection} \label{threshold-selection}
\label{app:threshold}
Our joint score ($\mathrm{Acc}_\mathrm{in} + R_\mathrm{oos}$) in Section~4.2 gives the same weight to the two metrics, $\mathrm{Acc}_\mathrm{in}$ and $R_\mathrm{oos}$, compared to other combined metrics like $(C_\mathrm{in}+C_\mathrm{oos})/(N_\mathrm{in}+N_\mathrm{oos})$.
Such a combined metric can put much more weight on the in-domain accuracy when $N_\mathrm{in}$ and $N_\mathrm{oos}$ are imbalanced; Table~2 shows such imbalance on the development set.
\citet{oos-intent} sacrificed the OOS recall a lot, and the trade-off with respect to the threshold selection was not discussed.

\section{Training Details} \label{training-details}

\begin{table*}[t]
\centering
\resizebox{1.0\linewidth}{!}{
\begin{tabular}{l|ccc|ccc}
\hline
           & \multicolumn{3}{c|}{\textbf{Single domain}}                 & \multicolumn{3}{c}{\textbf{All domains}}                 \\ \cline{2-7} 
           & \textbf{Learning rate} & \textbf{Epoch}    & \textbf{Run} & \textbf{Learning rate} & \textbf{Epoch} & \textbf{Run} \\ \hline
Classifier & \{1e-4, 2e-5, 5e-5\}   & \{15, 25, 35\}    & 10             & \{1e-4, 5e-5\}         & \{15, 25, 35\} & 5              \\
Emb-kNN    & \{1e-4, 2e-5, 3e-5\}   & \{7, 10, 20, 25, 35\} & 10             &   \{2e-5, 5e-5\}        &  \{3, 5, 7\}        & 5              \\
DNNC       & \{1e-5, 2e-5, 3e-5, 4e-5\}   & \{7, 10, 15\}     & 10             & \{2e-5, 5e-5\}         & \{3, 5, 7\}   & 5              \\ \hline
\end{tabular}}\caption{some hyper-parameter settings for a few models.}\label{table:hyper-paramter}
\end{table*}
%%%%%%%%%%%%%%%%%%%%%%%%%%%%%%%%%%%%%%%%%%%%%%%%%%%%%%%

%%%%%%%%%%%%%%%%%%%
\begin{table*}[t]
\centering
%\resizebox{\linewidth}{!}{
\begin{tabular}{l|cc}
\hline
           & \textbf{5-shot}                 & \textbf{10-shot}                \\ \hline
Classifier & \{bs: 50, ep: 25.0, lr: 5e-05\} & \{bs: 50, ep: 35.0, lr: 5e-05\} \\ \hline
Emb-kNN    & \{bs: 200, ep: 7.0, lr: 2e-05\} & \{bs: 200, ep: 5.0, lr: 2e-05\}  \\ \hline
DNNC       & \{bs: 900, ep: 7.0, lr: 2e-05\} &  \{bs: 1800, ep: 5.0, lr: 2e-05\} \\ \hline
\end{tabular}
%}
\caption{Best hyper-parameter settings for a few models on the all-domain experiments, where {\tt bs} is batch size, {\tt ep} represents epochs, {\tt lr} is learning rate.}\label{table:hyper-paramter-best-all}
\end{table*}

\paragraph{Dataset preparation}
To use the CLINC150 dataset~\citep{oos-intent}\footnote{\url{https://github.com/clinc/oos-eval}.} in our ways, especially for the single-domain experiments, we provide a zip file {\tt data\_preprocess\_for\_emnlp2020.zip} accompanied with the paper submission.

\paragraph{General training}\label{appendix-general-training}
This section describes the details about the model training in Section~4.3.
For each component related to RoBERTa and SRoBERTa, we solely follow the two libraries, transformers and sentence-transformers, for the sake of easy reproduction of our experiments.\footnote{\url{https://github.com/huggingface/transformers} and \url{https://github.com/UKPLab/sentence-transformers}.}
The example code to train the NLI-style models is also available.\footnote{\url{https://github.com/huggingface/transformers/tree/master/examples/text-classification}.}
We use the {\tt roberta-base} configuration\footnote{\url{https://s3.amazonaws.com/models.huggingface.co/bert/roberta-base-config.json}.} for all the RoBERTa/SRoBERTa-based models in our experiments.
All the model parameters including the RoBERTa parameters are updated during all the fine-tuning processes, where we use the AdamW~\citep{adamw} optimizer with a weight decay coefficient of 0.01 for all the non-bias parameters.
We use a gradient clipping technique~\citep{clip} with a clipping value of 1.0, and also use a linear warmup learning-rate scheduling with a proportion of 0.1 with respect to the maximum number of training epochs. 

\paragraph{Pre-training on NLI tasks}\label{appendix-pre-training}
For the pre-training on NLI tasks, we fine-tune a {\tt roberta-base} model on three publicly available datasets, i.e., SNLI~\citep{snli}, MNLI~\citep{mnli}, and WNLI~\citep{wnli} from the GLUE benchmark~\citep{glue}.
The optimizer and gradient clipping follow the above configurations.
The number of training epochs is set to $4$; the batch size is set to $32$; the learning rate is set to $2e-5$.
We use a linear warmup learning-rate scheduling with a proportion of $0.06$ by following \citet{roberta}.
The evaluation results on the development sets are shown in Table~\ref{table-pretrain}, where the low accuracy of WNLI is mainly caused by the data size imbalance.
We note that these NLI scores are not comparable with existing NLI scores, because we converted the task to the binary classification task for our model transfer purpose.

\paragraph{Text pre-processing}
For all the RoBERTa-based models, we used the RoBERTa {\tt roberta-base}'s tokenizer provided in the transformers library.\footnote{\url{https://github.com/huggingface/transformers/blob/master/src/transformers/tokenization_roberta.py}.}
We did not perform any additional pre-processing in our experiments.

\paragraph{Hyper-parameter settings}\label{appendix-hyper-parameter}
Table~\ref{table:hyper-paramter} shows the hyper-parameters we tuned on the development sets in our RoBERTa-based experiments.
For a single-domain experiment, we take a hyper-parameter set and apply it to the ten different runs to select the threshold in Section~4.2 on the development set.
We then select the best hyper-parameter set along with the corresponding threshold, and finally apply the model and the threshold to the test set.
We follow the same process for the all-domain experiments, except that we run each experiment five times.
Table~\ref{table:hyper-paramter-best-all} and Table~\ref{table:hyper-paramter-best} summarize the hyper-parameter settings used for the evaluation on the test sets.
We note that each model was not very sensitive to the different hyper-parameter settings, as long as we have a large number of training iterations.
%The hyper-parameter settings and the best hyper-parameter settings based on the validation sets for few models are shown in Table \ref{table:hyper-paramter}, Table \ref{table:hyper-paramter-best} and Table \ref{table:hyper-paramter-best-all}.

\begin{table}[]
\centering
\resizebox{\linewidth}{!}{
\begin{tabular}{l|lll}
\hline
Dataset                    & SNLI & WNLI & MNLI \\ \hline
Size of the development set & 9999          & 70            & 9814          \\
Accuracy                   & 94.5\%        & 41.4\%        & 92.1\%        \\ \hline
\end{tabular}}\caption{Development results on three NLI datasets.}
\label{table-pretrain}
\end{table}

\begin{table*}[t]
\centering
\resizebox{\linewidth}{!}{
\begin{tabular}{l|cccc}
\hline
           & \textbf{5-shot}                                      & \multicolumn{1}{c|}{\textbf{10-shot}}                 & \textbf{5-shot}                  & \textbf{10-shot}                 \\ \cline{2-5} 
           & \multicolumn{2}{c|}{\textbf{Banking}}                                                                                 & \multicolumn{2}{c}{\textbf{Credit cards}}                                    \\ \hline
Classifier & \multicolumn{1}{l}{\{bs: 15, ep: 25.0, lr: 5e-05\}}  & \multicolumn{1}{l|}{\{bs: 15, ep: 35.0, lr: 5e-05\}}  & \{bs: 15, ep: 15.0, lr: 5e-05\}  & \{bs: 15, ep: 25.0, lr: 5e-05\}  \\
Emb-kNN    & \multicolumn{1}{l}{\{bs: 200, ep: 35.0, lr: 1e-05\}} & \multicolumn{1}{l|}{\{bs: 200, ep: 25.0, lr: 2e-05\}} & \{bs: 100, ep: 20.0, lr: 1e-05\} & \{bs: 100, ep: 10.0, lr: 1e-05\} \\
DNNC       & \multicolumn{1}{l}{\{bs: 370, ep: 15.0, lr: 1e-05\}} & \multicolumn{1}{l|}{\{bs: 370, ep: 7.0, lr: 2e-05\}}  & \{bs: 370, ep: 15.0, lr: 2e-05\} & \{bs: 370, ep: 7.0, lr: 3e-05\}  \\ \hline
           & \multicolumn{2}{c}{\textbf{Work}}                                                                            & \multicolumn{2}{c}{\textbf{Travel}}                                 \\ \hline
Classifier & \{bs: 15, ep: 15.0, lr: 5e-05\}                      & \{bs: 15, ep: 15.0, lr: 5e-05\}                       & \{bs: 15, ep: 35.0, lr: 5e-05\}  & \{bs: 15, ep: 25.0, lr: 1e-04\}  \\
Emb-kNN    & \{bs: 100, ep: 20.0, lr: 1e-05\}                     & \{bs: 100, ep: 7.0, lr: 2e-05\}                       & \{bs: 100, ep: 35.0, lr: 3e-05\} & \{bs: 100, ep: 20.0, lr: 1e-05\} \\
DNNC       & \{bs: 370, ep: 7.0, lr: 3e-05\}                      & \{bs: 370, ep: 15.0, lr: 2e-05\}                      & \{bs: 370, ep: 7.0, lr: 2e-05\}  & \{bs: 370, ep: 7.0, lr: 2e-05\}  \\ \hline
\end{tabular}}\caption{Best hyper-parameter settings for a few models on the four single domains, where {\tt bs} is batch size, {\tt ep} represents epochs, {\tt lr} is learning rate.}\label{table:hyper-paramter-best}
\end{table*}

\section{Data Augmentation} \label{data-augmentation}

We describe the details about the classifier baselines with the data augmentation techniques in Section~4.3.
%In this paper, we apply the ``Classifier-EDA''~\cite{eda} and ``Classifier-BT''~\citep{qanet,backtrans-da} as two data augmentation baselines.

\paragraph{EDA}
Classifier-EDA uses the following four data augmentation techniques in \citet{eda}: synonym replacement, random insertion, random swap, and random deletion.
We follow the publicly available code.\footnote{\url{https://github.com/jasonwei20/eda_nlp}.}
For every training example, we empirically set one augmentation based on every technique.
We apply each technique separately to the original sentence and therefore every training example will have four augmentations.
The probability of a word in an utterance being edited is set to 0.1 for all the techniques.   

\paragraph{BT}
For classifier-BT, we use the English-German corpus in \citet{escape}, which is widely used in an annual competition for automatic post-editing research on IT-domain text~\citep{ape-2019}.
The corpus contains about 7.5 million translation pairs, and we follow the {\it base} configuration to train a transformer model~\citep{transformer} for each direction.
Based on the initial trial in our preliminary experiments to generate diverse examples, we decided to use a temperature sampling technique instead of a greedy or beam-search strategy.
More specifically, logit vectors during the machine translation process are multiplied by $\tau$ to distort the output distributions, where we set $\tau = 5.0$.
For each training example in the intent detection dataset, we first translate it into German and then translate it back to English.
We repeat this process to generate up to five unique examples, and use them to train the classifier model.
Table~\ref{tb:bt-examples} shows such examples, and we will release all the augmented examples for future research.

\begin{table*}[t]
  \begin{center}
{\small
    \begin{tabular}{l|l|l}
    
    Original utterance & Augmented example & Intent label \\ \hline
    can you block my chase account right away please & can you turn my chase account off directly & freeze account \\
    do a car payment from my savings account & with my saving account, you can pay a car payment account & pay bill \\
    when is my visa due & when is my visa to be paid & bill due \\ \hline

    \end{tabular}
}
    \caption{Examples used to train clasifier-BT.}
    \label{tb:bt-examples}
  \end{center}

\end{table*}

\section{More Results}
\label{extra-results}

\paragraph{Visualization} \label{appendix-vidualization}
%Figure~\ref{fig:visulization-appendix} shows the 5-shot and 10-shot development results in the banking domain. In-domain accuracy and OOS precision for DNNC-scratch and Classifier drop quickly with the increase of threshold, while the OOS precision increases quickly in the meantime. DNNC demonstrates higher robustness across all metrics with changes in the threshold. The results in Figure~\ref{fig:tsne-appendix} also shows the effectiveness of DNNC.
Figure~\ref{fig:visulization-appendix} shows the same curves in Figure~3 along with the corresponding 10-shot results.
We can see that the 10-shot results also exhibit the same trend.
Figure~\ref{fig:tsne-appendix} shows more visualization results with respect to Figure~1.
Again, the 10-shot visualization shows the same trend.

Figure~\ref{fig:Conf-appendix} and Figure \ref{fig:Conf-appendix-all-domains} show 5-shot and 10-shot confidence levels on the test sets of the banking domain and all domains, respectively.
Both Classifier and Emb-kNN cannot perform well to distinguish the in-domain examples from the OOS examples, while DNNC has a clearer distinction between the two.

\paragraph{Faster inference}\label{appendix-DNNC-joint}
Figure~\ref{fig:joint_nli-appendix} shows the same curves in Figure~4 also for the 10-shot setting.
We can see the same trend with the 10-shot results.

\paragraph{Case studies}\label{Case Study}
Table~\ref{table:case-study} shows four DNNC prediction examples from the development set of the banking domain.
For the first example, the input utterance is correctly predicted with a high confidence score, and it has a similarly matched utterance to the input utterance;
for the second example, the input utterance is predicted incorrectly with a high confidence score, where the matched utterance is related to money but it has a slightly different meaning with the input utterance.
For the third example, the model gives a very low confidence score to predict an OOS user utterance as an in-domain intent; the last example is an incorrect case where the input utterance and the matched utterance have a topically similar meaning, resulting in a high confidence score for the wrong label, ``bill due.''
Based on these observations, it is an important direction to improve the model's robustness (even with the large-scale pre-trained models) towards such confusing cases.

\begin{figure*}[t]
	\begin{center}
    	\includegraphics[width=0.85\linewidth]{./images_appendix/Visualization_Metric_Appendix.png}
    \end{center}
\caption{5-shot and 10-shot development results on the banking domain. In this series of plots, a model with a higher area-under-the-curve is more robust.}
\label{fig:visulization-appendix}
\end{figure*}

\begin{figure*}[t]
	\begin{center}
    	\includegraphics[width=0.85\linewidth,height=0.9\textheight]{./images_appendix/TSNE_Appendix.png}
    \end{center}
\caption{5-shot and 10-shot tSNE visualizations on development set of the banking domain, where circles represent in-domain intent classes, and red stars represent out-of-scope intents.}
\label{fig:tsne-appendix}
\end{figure*}

\begin{figure*}[t]
	\begin{center}
    	\includegraphics[width=0.85\linewidth,keepaspectratio=true,height=0.95\textheight]{./images_appendix/Conf_Appendix_with_sbert.png}
    \end{center}
\caption{5-shot and 10-shot confidence levels on test set of the banking domain. Best viewed in color.}
\label{fig:Conf-appendix}
\end{figure*}

\begin{figure*}[t]
	\begin{center}
    	\includegraphics[width=0.85\linewidth]{./images_appendix/Conf_Appendix_All_domains.png}
    \end{center}
\caption{5-shot and 10-shot confidence levels on test set of all domains. Best viewed in color.}
\label{fig:Conf-appendix-all-domains}
\end{figure*}

\begin{figure*}[t]
	\begin{center}
    	\includegraphics[width=0.85\linewidth]{./images_appendix/Visualization_Joint_nli_Appendix.png}
    \end{center}
\caption{5-shot and 10-shot DNNC-joint development results on the banking domain, where the dash lines are DNNC results.}
\label{fig:joint_nli-appendix}
\end{figure*}

\begin{table*}[]
\centering
\resizebox{1.0\linewidth}{!}{
\begin{tabular}{ll}
\hline
\textbf{input utterance}    & transfer ten dollars from my wells fargo account to my bank of america account              \\
\textbf{matched utterance} & transfer \$10 from checking to savings                                                      \\
\textbf{label of the input utterance}      & transfer                                                                                    \\
\textbf{label of the matched utterance}      & transfer                                                                                    \\
\textbf{confidence score}             & 0.934                                                                                      \\ \hline
\textbf{input utterance}    & what transactions have i accrued buying dog food                                            \\
\textbf{matched utterance} & what have i spent on food recently                                                          \\
\textbf{label of the input utterance}      & transactions                                                                                \\
\textbf{label of the matched utterance}      & spending history                                                                           \\
\textbf{confidence score}             & 0.915                                                                                      \\ \hline
\textbf{input utterance}    & who has the best record in the nfl                                                          \\
\textbf{matched utterance} & do i have enough in my boa account for a new pair of skis                                   \\
\textbf{label of the input utterance}      & OOS                                                                                         \\
\textbf{label of the matched utterance}      & balance                                                                                     \\
\textbf{confidence score}             & 0.006                                                                                       \\ \hline
\textbf{input utterance}    & how long will it take me to pay off my card if i pay an extra \$50 a month over the minimum \\
\textbf{matched utterance} & how long do i have left to pay for my chase credit card                                     \\
\textbf{label of the input utterance}      & OOS                                                                                         \\
\textbf{label of the matched utterance}      & bill due                                                                                   \\
\textbf{confidence score}             & 0.945                                                                                      \\ \hline
\end{tabular}} \caption{Case studies on the development set of banking domain. The first two cases are in-domain examples from the banking domain, and the rest are OOS examples.}\label{table:case-study}
\end{table*}